\documentclass{article}
\usepackage{amsmath,epsfig}
\usepackage{url}
\usepackage[preprint]{spconfa4}

\copyrightnotice{978-1-6654-3864-3/21/\$31.00 ©2021 IEEE}

\let\OLDthebibliography\thebibliography
\renewcommand\thebibliography[1]{
  \OLDthebibliography{#1}
  \setlength{\parskip}{0pt}
  \setlength{\itemsep}{0pt plus 0.3ex}
}

\begin{document}\sloppy

\def\x{{\mathbf x}}
\def\L{{\cal L}}

\title{Rotation Transformation Network: Learning View-Invariant Point Cloud for Classification and Segmentation}
%
\name{Shuang Deng$^{\ast}$$^{\dagger}$$^{\ddagger}$, Bo Liu$^{\ast}$$^{\dagger}$$^{\ddagger}$, Qiulei Dong$^{\ast}$$^{\dagger}$$^{\ddagger}$, and Zhanyi Hu$^{\ast}$$^{\dagger}$
	\thanks{This work was supported by the National Natural Science Foundation of China (U1805264, 61991423), the Strategic Priority Research Program of the Chinese Academy of Sciences (XDB32050100), and the Open Research Fund from Key Laboratory of Intelligent Infrared Perception, Chinese Academy of Sciences. Corresponding author: Qiulei Dong.}}
\address{$^{\ast}$National Laboratory of Pattern Recognition, Institute of Automation, \\
	Chinese Academy of Sciences, China \\ 
	$^{\dagger}$School of Artificial Intelligence, University of Chinese Academy of Sciences, China \\
	$^{\ddagger}$Center for Excellence in Brain Science and Intelligence Technology, \\
	Chinese Academy of Sciences, China \\
	\{shuang.deng, qldong, huzy\}@nlpr.ia.ac.cn, liubo2017@ia.ac.cn}

\maketitle

\begin{abstract}
Many recent works show that a spatial manipulation module could boost the performances of deep neural networks (DNNs) for 3D point cloud analysis. In this paper, we aim to provide an insight into spatial manipulation modules. Firstly, we find that the smaller the rotational degree of freedom (RDF) of objects is, the more easily these objects are handled by these DNNs. Then, we investigate the effect of the popular T-Net module and find that it could not reduce the RDF of objects. Motivated by the above two issues, we propose a rotation transformation network for point cloud analysis, called RTN, which could reduce the RDF of input 3D objects to 0. The RTN could be seamlessly inserted into many existing DNNs for point cloud analysis. Extensive experimental results on 3D point cloud classification and segmentation tasks demonstrate that the proposed RTN could improve the performances of several state-of-the-art methods significantly.
\end{abstract}
\section{Introduction}
\label{sec:intro}

With the rapid development of 3D sensors, 3D point cloud analysis techniques have drawn increasing attention in recent years. Inspired by the great success of Deep Neural Networks (DNNs) in the image analysis filed, a large number of works~\cite{2017pointnet,2017pointnet++,2018spidercnn,2018dgcnn,2018pointcnn,2020grid} have utilized DNNs to handle various tasks in the field of 3D point cloud analysis.

It is generally believed that one of the main factors which impede the development of many existing DNNs for point cloud classification and segmentation is: the input sets of point clouds belonging to a same object category are generally view-dependent, and they undergo different rigid transformations (translations and rotations) relative to a unified view. Compared with translation transformations whose influences could be easily eliminated through coordinate centralization, rotation transformations are more difficult to be handled. Additionally, it still lacks experimental analysis of the influence of object poses on the performances of these DNNs in literature. It is noted that some recent works~\cite{2017pointnet,2018dgcnn} showed that a learnable module that allows spatial manipulation of data could significantly boost the performances of DNNs on various point cloud processing tasks, such as point cloud classification and segmentation. For example, the popular T-Net~\cite{2017pointnet,2018dgcnn} is a learnable module that predicts an transformation matrix with an orthogonal constraint for transforming all the input point clouds to a 3-dimensional latent canonical space and has significantly improved the performance of many existing DNNs. Despite its excellent performance, the poses of different point clouds transformed via T-Net are still up to some 3-degree-of-freedom rotations as analyzed in Section Methodology.

Motivated by the aforementioned issues, we firstly compare and analyze the influence of RDF of input 3D objects on several popular DNNs empirically in this paper, observing that the smaller the RDF of objects is, the better these DNNs consistently perform. This observation encourages us to further investigate how to reduce the RDF of objects via a learnable DNN module. Then, we evaluate the performances of the T-Net used in ~\cite{2017pointnet} and ~\cite{2018dgcnn}, and find that although it could manipulate 3D objects spatially and improve the DNNs' performances to some extent, it could not transform the input view-dependent data into view-invariant data with $0$ RDF in most cases. Finally, we propose a rotation transformation network, called RTN, which utilizes a Euler-angle-based rotation discretization manner to learn the pose of input 3D objects and then transforms them to a unified view. The proposed RTN has a two-stream architecture, where one stream is for global feature extraction while the other one is for local feature extraction, and we also design a self-supervised scheme to train the RTN.

In sum, our major contributions are three-fold:
\begin{itemize}
  \item We empirically verify that the smaller the RDF of objects is, the more easily these objects are handled by some state-of-the-art DNNs, and we find that the popular T-Net could not reduce the RDF of objects in most cases.
  \item To our best knowledge, the proposed RTN is the first attempt to learn the poses of 3D objects for point cloud analysis under a self-supervised manner. It could effectively transform view-dependent data to view-invariant data, and could be easily inserted into many existing DNNs to boost their performance on point cloud analysis.
  \item Extensive experimental results on point cloud classification and segmentation demonstrate that the proposed RTN could help several state-of-the-art methods improve their performances significantly.
\end{itemize}

\section{Related Work}

\subsection{Deep Learning for 3D Point Clouds}
PointNet~\cite{2017pointnet} is the pioneering method to directly process 3D point clouds using shared multi-layer perceptrons (MLPs) and max-pooling layers. 
PointNet++~\cite{2017pointnet++} extends PointNet by extracting multiple-scale features of local pattern. 
Spatial graph convolution based methods have also been applied to 3D point clouds. 
SpiderCNN~\cite{2018spidercnn} treats the convolutional kernel weights as a product of a simple step function and a Taylor polynomial. 
EdgeConv is proposed in DGCNN~\cite{2018dgcnn} where a channel-wise symmetric aggregation operation is applied to the edge features in both Euclidean and semantic spaces. 

\subsection{Rotation-Invariant Representation for 3D Point Clouds}
Rotation invariance is one of the most desired properties for object recognition. Addressing this issue, many existing works investigate how to learn rotation-invariant representations from the 3D point clouds. In ~\cite{2019shk,2019discrete,2019rotation, RaoL019, 2020prin}, different types of convolutional kernel are designed to directly extract approximately rotation-invariant features of the input 3D point clouds. In ~\cite{2019srinet,2019clusternet, 2019quatnn}, they propose to manually craft a strictly rotation-invariant representation in the input space and uses this representation to replace the 3D Euclidean coordinate as model input which will inevitably result in information loss. Unlike those above methods, this paper aims to learn a spatial transformation which transforms the input view-dependent 3D objects into view-invariant objects with 0 RDF.

\section{Methodology}

In this section, we firstly compare and analyze the influences of the rotational degree of freedom (RDF) of objects on the performances of four popular DNNs for point cloud analysis. Secondly, we investigate whether T-Net~\cite{2015stn} could reduce the RDF of objects or not. Finally, we describe the proposed rotation transformation network (RTN) in detail.

\begin{table}[t]
	\begin{center}
		\caption{Classification performances of four methods on 3D point clouds with different rotational degrees of freedom.}
		\label{tab_pre_modelnet}
		\resizebox{1\columnwidth}{!}{
			\begin{tabular}{|c|c|c|c|}
				\hline
				Method       &SO(0)(Ins/mCls)       &SO(1)(Ins/mCls)     &SO(3)(Ins/mCls)     \\
				\hline
				PointNet~\cite{2017pointnet}      &89.1/85.9      &88.1/85.2      &84.4/79.9   \\
				PointNet++~\cite{2017pointnet++}  &90.6/86.8      &89.9/86.2      &85.7/80.6   \\
				DGCNN~\cite{2018dgcnn}       	  &92.4/90.2  	 &91.4/88.8   	&88.7/84.4   \\
				SpiderCNN~\cite{2018spidercnn}    &91.5/87.8  	 &90.2/87.8   	&83.9/78.7   \\
				\hline
			\end{tabular}
		}
	\end{center}
\end{table}

\begin{figure}[t]
	\centering
	\includegraphics[width=1 \columnwidth]{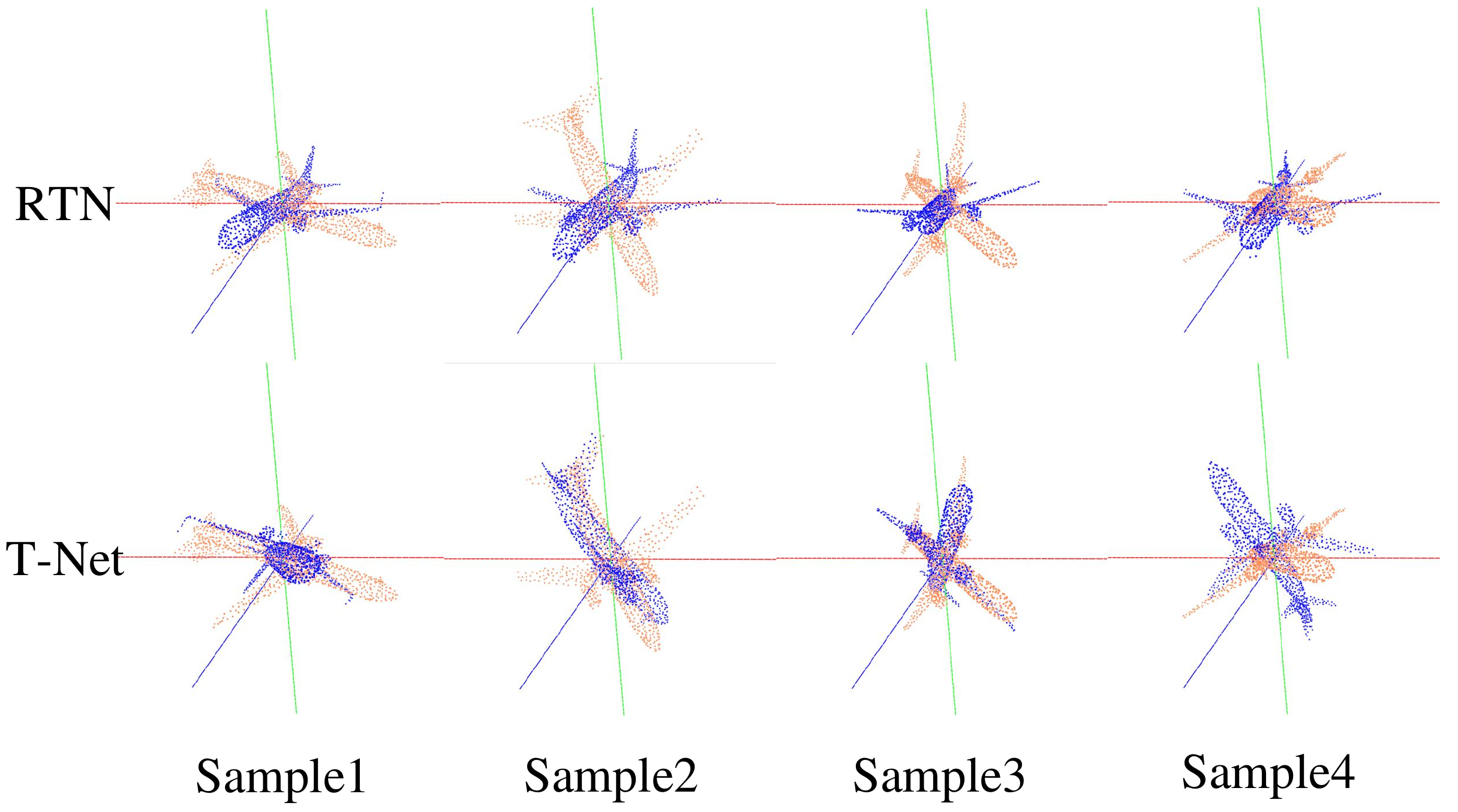}
	\caption{Visualization of point clouds before and after two spatial manipulation module (RTN and T-Net). The first line presents the results of RTN and the second line presents those of T-Net. The orange ones represent the point clouds before spatial manipulation while the blue ones represent those after spatial manipulation.}
	\label{fig_rotation}
\end{figure}

\begin{figure*}[t]
	\centering
	\includegraphics[width=1.95 \columnwidth]{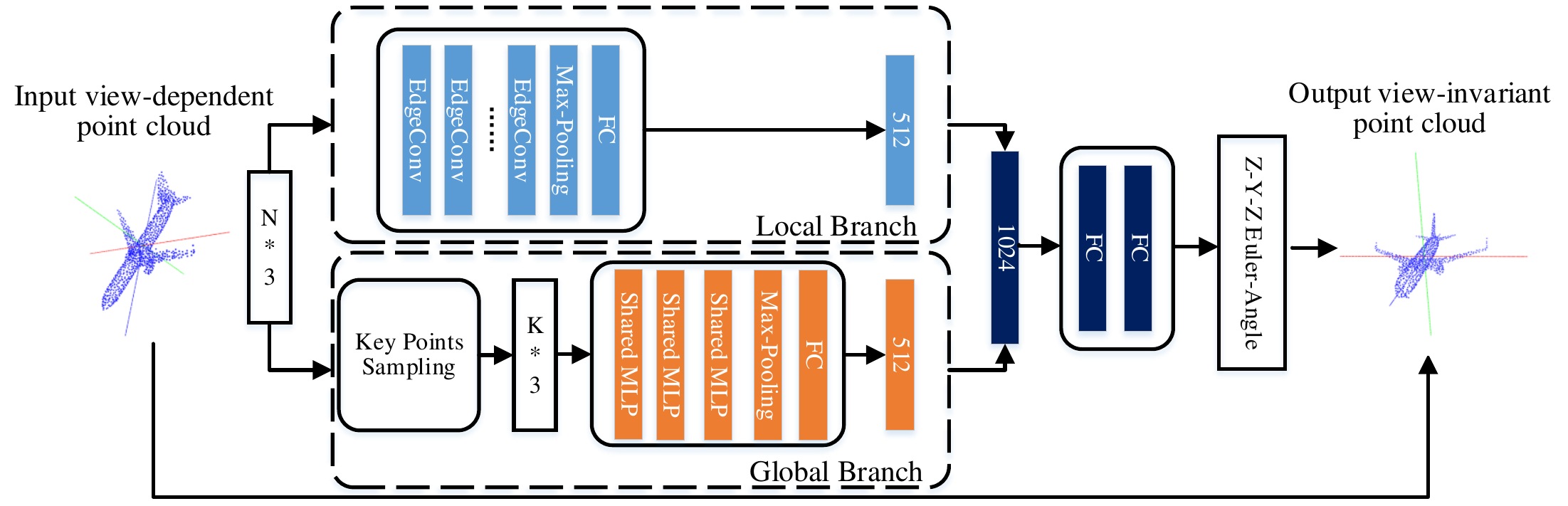}
	\caption{Architecture of the Proposed RTN.}
	\label{fig_architecture}
\end{figure*}

\subsection{Influences of RDF of Objects on DNNs}
We investigate the influences of the RDF of objects on four state-of-the-art methods including PointNet~\cite{2017pointnet}, PointNet++~\cite{2017pointnet++}, DGCNN~\cite{2018dgcnn}, and SpiderCNN~\cite{2018spidercnn}, where no special modules are employed for explicitly extracting rotation-invariant representations from 3D point clouds. These methods are trained and evaluated on point cloud classification with the following three sets of data:
\begin{itemize}
	\item Data SO(0): for the input objects belonging to each category, they locate a same pose in a centralized 3D space. The RDF of these objects is $0$.
	\item Data SO(1): for the input objects belonging to each category, they locate on a reference plane in a centralized 3D space. The RDF of these objects is $1$.
	\item Data SO(3): for the input objects belonging to each category, they locate with an arbitrary pose in a centralized 3D space. The RDF of these objects is $3$.
\end{itemize}
The instance accuracy (Ins(\%)) and average per-class accuracy (mCls(\%)) for the classification task on the public ModelNet40 dataset by the four methods are reported in Table \ref{tab_pre_modelnet}. We also investigate the influences of the RDF of objects on ShapenetPart for point cloud segmentation, which refers to the supplementary material. As seen from Table \ref{tab_pre_modelnet}, the classification performances by the referred methods on Data SO(0) and Data SO(1) are significantly higher than those on Data SO(3), and their performances on Data SO(0) are best in most cases. This demonstrates that the smaller the RDF of objects is, the more easily these objects are handled, which encourages us to investigate whether the popular T-Net used in some state-of-the-art methods~\cite{2017pointnet,2018dgcnn} could reduce the RDF of objects and how to design a more effective DNN module to do so in the following two subsections respectively.

\subsection{Could T-Net Reduce the RDF of Objects?}
The observation in the above subsection naturally raises the following question: Could the T-Net extensively used in some state-of-the-art methods~\cite{2017pointnet,2018dgcnn} reduce the RDF of objects or not? In theory, T-Net aims to learn a spatial transformation matrix with only an orthogonal constraint, and the learnt orthogonal matrix by T-Net could not strictly guarantee that the input view-dependent objects could be transformed into a unified view.

In order to further investigate the above question, we visualize many samples from each category in ModelNet40 and the corresponding transformed point clouds by the T-Net used in~\cite{2018dgcnn} \footnote{Due to the fact that the T-Net were used similarly in ~\cite{2017pointnet,2018dgcnn}, we only visualize the prediction results by the T-Net used in ~\cite{2018dgcnn}.}. Due to the limited space, the second row of Figure \ref{fig_rotation} shows four samples of the planes where the orange point clouds with $3$ RDF are the inputs to T-Net, while the blue point clouds are the corresponding transformed ones by T-Net. As seen in Figure \ref{fig_rotation}, the transformed point clouds by T-Net still have $3$ RDF. This demonstrates that T-Net could not reduce the RDF of objects.

\subsection{Rotation Transformation Network}
Inspired by the above observations, we investigate how to design a network for reducing the RDF of input object point clouds effectively. Here, we propose a rotation transformation network (RTN), which could learn the rotations of the input 3D objects and then use the learnt rotations to obtain view-invariant objects by performing inverse rotations. The architecture of the proposed RTN is shown in Figure \ref{fig_architecture}. 

In the proposed RTN, the rotation learning problem is transformed into a classification problem where a Euler-angle-based rotation discretization is employed. Then a self-supervised learning scheme is designed to train the proposed RTN. In the following, we firstly give a detailed explanation on the Euler-angle-based rotation discretization in our network. Then, we describe the detailed architecture. Lastly, we present the details of the proposed self-supervised learning scheme. 

\textbf{3D Rotation Discretization.} 
Here, our goal is to discretize infinite 3-degree-of-freedom rotations into a finite group of rotation classes. We use the Z-Y-Z Euler-angle representation under a world coordinate system: An arbitrary 3D rotation is accomplished by firstly rotating the object around the Z axis by angle $\alpha$, and secondly rotating it around the Y axis by angle $\beta$, and lastly rotating it around the Z axis by angle $\gamma$, which is also formulated by the following equation:
\begin{eqnarray}
	\begin{split}
		&\boldsymbol{R}(\alpha,\beta,\gamma)=\boldsymbol{R}_{Z}(\gamma) \circ \boldsymbol{R}_{Y}(\beta) \circ \boldsymbol{R}_{Z}(\alpha) \\
		&\textup{s.t.} \quad \alpha \in [0,2\pi), \beta \in [0,\pi], \gamma \in [0,2\pi) \\
	\end{split}
	\label{equ1}
\end{eqnarray}
where $\boldsymbol{R}(\alpha,\beta,\gamma)$ indicates an arbitrary 3D rotation, $\boldsymbol{R}_{Z}(\alpha)$ $($also $\boldsymbol{R}_{Z}(\gamma)$$)$ indicates a rotation with $\alpha$ $($also $\gamma$$)$ around the Z axis, $\boldsymbol{R}_{Y}(\beta)$ indicates a rotation with $\beta$ around the Y axis, and $\circ$ means matrix multiplication.

After defining the Z-Y-Z Euler-angle representation of 3D rotations, we discretize the continuous range of \{$\alpha$, $\beta$, $\gamma$\} into a set of discrete values. In detail, we uniformly discretize the range of $\alpha \in [0,2\pi)$ into $N_1 = 2\pi/\theta$ values with a pre-fixed interval $\theta$. To avoid singular points, we adopt a sphere equiangular discretization to jointly discretize $\beta \in [0,\pi]$ and $\gamma \in [0,2\pi)$ into $N_2=(\frac{\pi}{\theta}-1) \times \frac{2\pi}{\theta} + 2$ values with interval $\theta$. Then, the total number of rotation classes is $N = N_1 \times N_2$. Note that the discretized rotation classes will become more fine-grained (larger $N$) as the interval $\theta$ becomes smaller.

\textbf{Network Architecture.}
As shown in Figure \ref{fig_architecture}, the proposed RTN employs a global branch and a local branch, where the local branch uses local aggregation method to extract features and the global branch only extracts point-wise features of the key points. The inputs to the RTN are point clouds with an arbitrary view, while its outputs are the corresponding view-invariant point clouds.

The global branch firstly samples $M$ key points of the 3D objects, which is described in the supplementary material specifically. Then, these $M$ key points are used to extract the point-wise features via three shared MLP layers, and a max-pooling followed by a fully-connected layer is applied to the features of these key points.

The local branch takes dense points clouds as inputs and employs five EdgeConv~\cite{2018dgcnn} layers to extract features. The last EdgeConv layer takes as input the feature concatenated by the outputs of the preceding EdgeConv layers to aggregate local features of the point clouds, and the final feature is obtained by a max-pooling layer followed by a fully-connected layer.

After obtaining the features from the global branch and the local branch, we concatenate and feed them into fully-connected layers to predict a discretized rotation class. Once the rotation of an input object relative to the unified view is obtained, an inverse rotation is applied to the input object to obtain its corresponding view-invariant point cloud.

\textbf{Self-Supervised Rotation Training.} 
Here, a self-supervised scheme for generating labeled training samples is introduced. Assuming that some samples with a fixed view are given, for each sample, we firstly generate a random Z-Y-Z Euler-angle-based rotation. Its rotation label $y$ is obtained according to the discretized \{$\alpha,\beta,\gamma$\} rotation angles, where $y \in \left\{1,2,\cdots, N\right\}$ and $N$ is the number of all classes of discretized rotations. Then we apply the generated 3D rotation to the sample under a world coordinate system for generating a new sample. Accordingly, we could obtain a large amount of labeled samples with different views and utilize them to train the RTN via multi-class cross-entropy loss.

\section{Experiments}

In this section, we firstly introduce the experimental setup. Secondly, we evaluate the rotation estimation performance of the proposed RTN. Then we give the comparative experimental results on the classification and segmentation tasks. Lastly, we end up with ablation analysis. Additionly, we also provide experiments on the effect of different rotation representations in the supplementary material. The code will be available at \url{https://github.com/ds0529/RTN}.

\subsection{Experimental Setup}
We evaluate the proposed method on the ModelNet40 shape classification benchmark~\cite{2015modelnet} and the ShapenetPart part segmentation benchmark~\cite{2015shapenet}. The poses of shapes in ModelNet40 is not totally aligned, so we manually rotated the shapes belonging to an same category to locate at an same pose for precise alignment. The pose of all the shapes in ShapenetPart is aligned precisely. The discretization interval of $R(\alpha,\beta,\gamma)$ is set to $\pi/6$, so that $N$ is $744$. The details of datasets and network parameters are described in the supplementary material.

\begin{table}[t]
	\begin{center}
		\caption{The mean inCD and outCD values of RTN on ModelNet40 and ShapenetPart.}
		\label{tab_rotation1}		
		\vspace{1em}\resizebox{0.8\columnwidth}{!}{
			\begin{tabular}{|c|c|c|}
				\hline
				Dataset     &ModelNet40    &ShapenetPart\\
				\hline
				Mean inCD                     & 0.19   & 0.21  \\
				Mean outCD					 & 0.09   & 0.08  \\
				\hline
			\end{tabular}
		}
	\end{center}
\end{table}

\subsection{Performance of RTN on Rotation Estimation}
We evaluate the rotation estimation performance of the proposed RTN on ModelNet40 and ShapenetPart through Chamfer Distance (CD)~\cite{2017pgn} and rotation classification accuracy. CD can directly evaluate the quality of rotation estimation but the other can not due to symmetric 3D objects. The details of rotation classification results are discribed in the supplementary material. 

CD calculates the average closest point distance between two point clouds. For each 3D object, we calculate two CD values, one of which is between input rotated point cloud and the point cloud with $0$ RDF (inCD), and the other is between output point cloud by proposed RTN and the point cloud with $0$ RDF (outCD). Then we average the calculated CD values of all 3D objects. We perform the experiments five times independently and use the mean results as the final results. 

The mean CD values are listed in Table \ref{tab_rotation1}. As seen from Table \ref{tab_rotation1}, the mean outCD values on both datasets are pretty lower than the mean inCD values, which indicates that the proposed RTN has ability to transform the input 3-RDF point clouds to 0-RDF point clouds in most cases. Furthermore, we visualize the input rotated point clouds in ModelNet40 and the corrected counterpart via two spatial manipulation module (RTN and T-Net~\cite{2018dgcnn}) in Figure \ref{fig_rotation}. The visualization shows that T-Net could not reduce the RDF of objects, but the proposed RTN could effectively reduce $3$ RDF from them.

\begin{table}[t]
	\begin{center}
		\caption{Comparison on ModelNet40 with Data SO(3) for 3D point cloud classification. }
		\label{tab_modelnet}
		\resizebox{1\columnwidth}{!}{
			\begin{tabular}{|c|c|c|c|}
				\hline
				Method                           & Input(size) 				& Ins/mCls       \\
				\hline
				PointNet(with T-Net)~\cite{2017pointnet} $\diamondsuit$     & pc(1024$\times$3)  		&84.4/79.9           \\
				PointNet++~\cite{2017pointnet++} $\diamondsuit$ & pc(1024$\times$3)  	    &85.7/80.6          \\
				DGCNN(with T-Net)~\cite{2018dgcnn} $\diamondsuit$          & pc(1024$\times$3)  	    &88.7/84.4          \\
				SpiderCNN~\cite{2018spidercnn} $\diamondsuit$   & pc(1024$\times$3)  		&84.0/78.7    		     \\
				\hline
				Zhang et al.\cite{2019rotation} $\heartsuit$           & pc(1024$\times$3) 		&86.4/-          \\
				Poulenard et al.\cite{2019shk} $\heartsuit$                  & pc(1024$\times$3) 		&87.6/-    		   \\
				Li et al.\cite{2019discrete} $\heartsuit$             & pc+normal(1024$\times$6) 		&88.8/-    		   \\
				ClusterNet~\cite{2019clusternet} $\heartsuit$ & pc(1024$\times$3)        &87.1/-    		  \\
				SRINet~\cite{2019srinet} $\heartsuit$        & pc+normal(1024$\times$6)        &87.0/-    		 \\
				REQNNs~\cite{2019quatnn} $\heartsuit$        & pc(1024$\times$3)        &83.0/-    		 \\
				\hline
				Ours(RTN+PointNet)    				                     & pc(1024$\times$3)        & 86.0/81.0          \\
				Ours(RTN+PointNet++)    				                     & pc(1024$\times$3)        & 87.4/82.7          \\
				Ours(RTN+DGCNN)    				                     & pc(1024$\times$3)        & \textbf{90.2}/\textbf{86.5}          \\
				Ours(RTN+SpiderCNN)    				                     & pc(1024$\times$3)        & 86.6/82.4          \\
				\hline
			\end{tabular}
		}
	\end{center}
\end{table}

\subsection{3D Point Cloud Classification}
Here, we combine the proposed RTN with four state-of-the-art methods including PointNet~\cite{2017pointnet}, PointNet++~\cite{2017pointnet++}, DGCNN~\cite{2018dgcnn}, and SpiderCNN~\cite{2018spidercnn} respectively, denoted as RTN+PointNet, RTN+PointNet++, RTN+DGCNN, and RTN+SpiderCNN, and evaluate their performances on 3D point cloud classification task. The models are trained and tested with Data SO(3) on ModelNet40 for comparing the performance on 3D rotation invariance, and two criteria are used to evaluate the performance: instance accuracy (denoted as Ins (\%)) and average per-class accuracy (denoted as mCls (\%)). We perform the experiments five times independently and use the mean results as the final results. We compare the results of the proposed methods with nine recent state-of-the-art methods as summarized in Table \ref{tab_modelnet}. In Table \ref{tab_modelnet}, the results of the four methods marked by $\diamondsuit$ are obtained by re-implementing these methods by the authors, because these methods are not evaluated on Data SO(3) in the original papers, while the results of the five methods marked by $\heartsuit$ are cited from the original papers directly. As noted from Table \ref{tab_modelnet}, we find that the proposed RTN is able to help the existing DNNs to improve their performances on dealing with 3D rotation variance by transforming the input view-dependent point clouds to view-invariant point clouds. The comparative results also show us that the RTN-based DNNs are superior to the T-Net-based DNNs, which informs us that the proposed RTN is better at reducing RDF than T-Net. The DGCNN equipped with the proposed RTN outperforms the current state-of-the-art methods with significant improvement.

\begin{table}[t]
	\begin{center}
		\caption{Comparison on ShapenetPart with Data SO(3) for 3D point cloud segmentation.}
		\label{tab_shapenet1}
		\vspace{1em}\resizebox{1\columnwidth}{!}{
			\begin{tabular}{|c|c|c|}
				\hline
				Method     						 & Input(size) 				& mIoU/Acc            \\
				\hline
				PointNet(with T-Net)~\cite{2017pointnet} $\diamondsuit$     & pc(2048$\times$3)  		& 79.1/90.6    	     \\
				PointNet++~\cite{2017pointnet++} $\diamondsuit$ & pc(2048$\times$3) & 75.4/88.4          \\
				DGCNN(with T-Net)~\cite{2018dgcnn} $\diamondsuit$      	 & pc(2048$\times$3)  	    & 78.9/90.8          \\
				SpiderCNN~\cite{2018spidercnn} $\diamondsuit$   & pc(2048$\times$3) & 74.5/87.9         		 \\
				\hline
				Zhang et al.\cite{2019rotation} $\heartsuit$  			 & pc(2048$\times$3) 		& 75.5/-             \\
				SRINet~\cite{2019srinet} $\heartsuit$        & pc+normal(2048$\times$6)        & 77.0/89.2          \\
				\hline
				Ours(RTN+PointNet)    				                     & pc(2048$\times$3)        & 80.1/91.2         \\
				Ours(RTN+PointNet++)    				                     & pc(2048$\times$3)        & 80.0/91.0          \\
				Ours(RTN+DGCNN)    				                     & pc(2048$\times$3)        & \textbf{82.8}/\textbf{92.6}          \\
				Ours(RTN+SpiderCNN)    				                     & pc(2048$\times$3)        & 80.1/90.7         \\
				\hline
			\end{tabular}
		}
	\end{center}
\end{table}

\subsection{3D Point Cloud Segmentation}
Although the results in the classification task have demonstrated the effectiveness of the proposed RTN, we further evaluate the proposed RTN by conducting experiments in 3D point cloud segmentation task. We perform segmentation on ShapenetPart, and average per-shape IoU (denoted as mIoU (\%)) and point-level classification accuracy (denoted as Acc (\%)) are used to evaluate the performances. We also perform the experiments five times independently and use the mean results as the final results, where the models are trained and tested with Data SO(3). The results are compared with six recent state-of-the-art methods as listed in Table \ref{tab_shapenet1}. A more detailed comparison among the RTN based DNNs and the comparative methods is described in the supplementary material. As seen in Table \ref{tab_shapenet1}, the methods equipped with RTN lead to a significant improvement compared to the corresponding original methods without RTN respectively. The DGCNN equipped with the proposed RTN outperforms all the current methods. 

\begin{table}[t]
	\begin{center}
		\caption{Results of RTNs using different backbones on ModelNet40 with Data SO(3). GA means global architecture. LA means local architecture. GLA means global-local architecture.}
		\label{tab_architecture}
		\vspace{1em}\resizebox{0.6\columnwidth}{!}{
			\begin{tabular}{|c|c|c|c|}
				\hline
				Backbone     & GA		& LA	   &GLA     \\
				\hline
				Ins                      & 89.7   & 89.6  &\textbf{90.2}  \\
				mCls						 & 85.1   & 85.8  &\textbf{86.5}   \\
				\hline
			\end{tabular}
		}
	\end{center}
\end{table}

\begin{table}[t]
	\begin{center}
		\caption{Results of RTNs with different quantization intervals on ModelNet40 with Data SO(3).}
		\label{tab_interval}
		\vspace{1em}\resizebox{0.8\columnwidth}{!}{
			\begin{tabular}{|c|c|c|c|c|}
				\hline
				Quantization Interval     & $\pi/9$		& $\pi/6$	&$\pi/4$    &$\pi/3$ \\
				\hline
				Ins                      & 89.7   & \textbf{90.2}  &89.8   &89.5\\
				mCls						 & 86.0   & \textbf{86.5}  &85.9   &85.2\\
				\hline
			\end{tabular}
		}
	\end{center}
\end{table}

\subsection{Ablation Analysis}
\textbf{Effect of backbone.} To prove the superiority of the proposed global-local architecture(GLA), we perform the classification task on ModelNet40 with RTNs with the global architecture(GA), the local architecture(LA) and the global-local architecture. DGCNN is used as the classification network after RTN. The results under different backbone configurations are summarized in Table \ref{tab_architecture}. It shows that the proposed global-local architecture achieves the best performance among all the backbone configurations, which demonstrates the benefit of the global-local architecture.

\textbf{Effect of Discretization Interval.} The interval affects the rotation classification performance of RTN, and thus affects the performance of existing DNNs equipped with RTN for point cloud analysis. Here we conduct experiments to analyze the effect of the discretization interval by setting a group of intervals $\{\pi/9,\pi/6,\pi/4,\pi/3\}$ in the classification task on ModelNet40. The results are listed in Table \ref{tab_interval}. As seen from Table \ref{tab_interval}, the classification accuracies under the above internals are quite close, demonstrating that the proposed method is not sensitive to the angle interval. The interval $\pi/6$ achieves the best performanceand, so we use this interval in both classification and segmentation experiments.

\section{Conclusion}
In this paper, we firstly find that the smaller the RDF of objects is, the more easily these objects are handled by these DNNs. Then, we find that T-Net module has limited effect on reducing the RDF of input 3D objects. Motivated by the above two issues, we propose a rotation transformation network, called RTN, which has the ability to explicitly transform input view-dependent point clouds to view-invariant point clouds by learning the rotation transformation based on an Euler-angle-based rotation discretization manner. Extensive experimental results indicate that the proposed RTN is able to help existing DNNs significantly improve their performances on point cloud classification and segmentation.

\bibliographystyle{IEEEbib}
\bibliography{rtn}

\end{document}